\pdfoutput=1
\documentclass[11pt]{article}

\usepackage{authblk}
\usepackage{acl}
\usepackage{times}
\usepackage{latexsym}
\usepackage{graphicx}
\usepackage[T1]{fontenc}
\usepackage[utf8]{inputenc}
\usepackage{microtype}
\usepackage{smartdiagram}
\usesmartdiagramlibrary{additions}
\usepackage[shortlabels]{enumitem}
\usepackage{multirow}
\usepackage{subcaption}
\title{Sociocultural Considerations in Monitoring Anti-LGBTQ+ Content on Social Media}

\author[1,2,3]{Sidney G.-J. Wong}
\affil[1]{University of Canterbury, New Zealand}
\affil[2]{Geospatial Research Institute, New Zealand}
\affil[3]{New Zealand Institute of Language, Brain and Behaviour, New Zealand}
\affil[ ]{\texttt{\{sidney.wong\}@pg.canterbury.ac.nz}}

\begin{document}

\maketitle

\begin{abstract}
    The purpose of this paper is to ascertain the influence of sociocultural factors (i.e., social, cultural, and political) in the development of hate speech detection systems. We set out to investigate the suitability of using open-source training data to monitor levels of anti-LGBTQ+ content on social media across different national-varieties of English. Our findings suggests the social and cultural alignment of open-source hate speech data sets influences the predicted outputs. Furthermore, the keyword-search approach of anti-LGBTQ+ slurs in the development of open-source training data encourages detection models to overfit on slurs; therefore, anti-LGBTQ+ content may go undetected. We recommend combining empirical outputs with qualitative insights to ensure these systems are fit for purpose.
\end{abstract}

    \textbf{Content Warning}: This paper contains unobfuscated examples of slurs, hate speech, and offensive language with reference to homophobia and transphobia which may cause distress.

\section{Introduction}
\label{sec:introduction}

    The proliferation of hate speech on social media platforms continues to negatively impact LGBTQ+ communities \cite{stefania_hate_2021}. As a consequence of anti-LGBTQ+ hate speech, these already minoritised and marginalised communities may experience digital exclusion and barriers to access in the form of the digital divide \cite{norris_digital_2001}. There have been considerable developments within the field of Natural Language Processing (NLP) in response to this social issue \cite{sanchez-sanchez_mapping_2024}, with most of the methodological advancements in this area being made in the last three decades \cite{tontodimamma_thirty_2021}.
    
    While much of hate speech research has focused on documentation and detection, there has been little attention on how these approaches can be applied across different social, political, or linguistic contexts \cite{locatelli_cross-lingual_2023}. Just as the appropriateness of swear words is highly contextually variable depending on language and culture \cite{jay_pragmatics_2008}, hate speech in the form of anti-LGBTQ+ hate speech is often predicated by social, cultural, and political attitudes towards diverse gender and sexualities. With minimal literature beyond just a system development context, we set out to investigate the suitability of implementing open-source anti-LGBTQ+ hate speech system on real-world sources of social media data.
    
    This paper makes two contributions: firstly, we show the predicted outputs from classification models can be transformed into various time series data sets to monitor the rate and volume of anti-LGBTQ+ hate speech on social media. Secondly, we argue that social, cultural, and linguistic bias introduced during the data collection phase has an impact on the suitability of these approaches.

\subsection{Related Work}
\label{subsec:related_work}
    
    Hate speech detection is often treated as a text classification task, whereby existing data can be used to train machine learning models to predict the attributes of unknown data \cite{jahan_systematic_2023}. The main focus of these systems are racism, sexism and gender discrimination, and violent radicalism \cite{sanchez-sanchez_mapping_2024}. Both the production and deployment of hate speech detection systems are methodologically similar produced under the following pipeline \cite{kowsari_text_2019}: 
    
    \begin{enumerate}
    
        \item[a)] \textit{Data Set Collection and Preparation}: involves collecting either real-world or synthetic instances of hate speech in a language condition (i.e., keyword search). This phase may involve or manual annotation from experts of crowd-sourced annotators.
        
        \item[b)] \textit{Feature Engineering}: involves manipulating and transforming instances of hate speech. This may involve anonymisation or confidentialisation depending on the privacy and data use rules for each social media platform.
        
        \item[c)] \textit{Model Training}: involves developing a hate speech detection system with machine learning algorithms. This may involve statistical language models or incorporating transformer-based large language models.
        
        \item[d)] \textit{Model Evaluation}: involves producing model performance metrics to determine the statistical validity of the system. This may involve making predictions on unseen or test data.
        
    \end{enumerate}
    
    Despite their straightforward workflow, these systems pose a number of ethical challenges and risks to the vulnerable communities \cite{vidgen_directions_2020}. Cultural biases and harms can be introduced at each stage of the data set production process \cite{sap_risk_2019}. Some of this can be attributed to poorly designed systems which are not fit for purpose \cite{vidgen_directions_2020}. For example, racial bias was identified in one open-source hate speech detection system developed by \citet{davidson_automated_2017} which resulted in samples of written African American English being misclassified as instances of hate speech and offensive language \cite{davidson_racial_2019}. 
    
    The presence of racial bias can be attributed to the decisions made during the \textit{Data Set Collection and Preparation} phase during system development.  \citet{davidson_automated_2017} took a keyword search approach (i.e., slurs and profanities) to identify instances of hate speech and offensive language. These samples were then used in the development of the detection system. Although slurs and profanities are good evidence of anti-social behaviour, the same words can also be re-appropriated or reclaimed by target communities \cite{popa-wyatt_reclamation_2020}. Classifications algorithms are unable to account for implicit world knowledge.

    Similarly, simple machine learning algorithms cannot account for linguistic variation which is another form of implicit world knowledge. Of interest to our current investigation, \citet{wong_monitoring_2023} applied the same system developed by \citet{davidson_automated_2017} on samples of tweets/posts originating in New Zealand. The system erroneously classified tweets/posts with words such as \textit{bugger}, \textit{digger}, and \textit{stagger} as instances of hate speech. An unintended consequence of these misclassified tweets/posts is that rural areas exhibited higher rates of hate speech and offensive language when compared to the national mean.

    However, not all forms of biases stem from decisions made during system development. Recent innovations in transformer-based language models, such as \textsc{bert} \cite{devlin_bert_2019}, have introduced new ethical challenges as the presence of gender, race, and other forms of bias have been observed in the word embeddings of large language models \cite{tan_assessing_2019}. This means there is potential for bias even in the later stages of system development during the \textit{Model Training} phase.

    While we grow increasingly aware of the impacts from these limitations \cite{alonso_alemany_bias_2023}, the number of hate speech detection data sets and systems continue to increase \cite{tontodimamma_thirty_2021}. A systematic review of hate speech literature has identified over 69 training data sets to detect hate speech on online and social media for 21 different language conditions \cite{jahan_systematic_2023}. Seemingly, the solution to addressing social, cultural, and political discrepancies within hate speech detection is to develop more systems in different languages.

    There remains little interest from NLP researchers to consider the issue of hate speech detection from a social impact lens \cite{hovy_social_2016}. The primary concerns in this research area are largely methodological. For example, improving model performance of detection systems resulting from noisy training data \citep{arango_hate_2022}. \citet{laaksonen_datafication_2020} critiqued the \textit{datafication} of hate speech detection which in turn has become an unnecessary distraction for NLP researchers in combating this social issue. 

    In fact, the appetite in applying NLP approaches for social good has decreased over time \cite{fortuna_cartography_2021}. Some researchers are beginning to question whether the efforts put towards the development and production of hate speech detection systems is the ideal solution for this social issue \cite{parker_is_2023}. In sidelining these pressing issues in hate speech detection research, we may unintentionally perpetuate existing prejudices against marginalised and minoritised groups these systems were meant to support \cite{buhmann_towards_2021}.
    
    In light of these ethical and methodological challenges in hate speech detection \cite{das_toward_2023}, we are starting to see how sociolinguistic information can be used to fine tune and improve the social and cultural performance of hate speech detection (\citealp{wong_cantnlplt-edi-2023_2023}; \citealp{wong_cantnlplt-edi-2024_2024}) using well-attested methods such as domain adaptation \cite{liu_roberta_2019}. NLP researchers may still play an invaluable role in combating online hate speech by incorporating sociocultural considerations in the development and deployment of hate speech detection systems.
    
\section{Methodology}
\label{sec:methodology}

    \begin{table}[t]
        \centering
            \begin{tabular}{lccc}
            \hline \textbf{Hostility} & \textbf{Direct} & \textbf{Indirect} & \textbf{Total} \\ \hline
            Abusive & 20 & 45 & 65 \\
            Disrespectful & 5 & 56 & 61 \\
            Fearful & 5 & 47 & 52 \\
            Hateful & 36 & 106 & 142 \\
            Normal & 13 & 71 & 84 \\
            Offensive & 65 & 308 & 373 \\
            \textbf{Total} & 144 & 633 & 777 \\
            \hline
        \end{tabular}
        \caption{\label{tab:ousidhoum} The distribution of English posts/tweets and the level of hostility by directness targeting sexual orientation in \citet{ousidhoum_multilingual_2019}. Note that all totals are total responses.}
    \end{table}
    
    As discussed in Section \ref{subsec:related_work}, hate speech detection research needs to undergo a paradigmatic shift in order to truly enable positive social impact, social good, and social benefit potential. The main purpose of this paper is to ascertain the influence of sociocultural factors (i.e., social, cultural, and political) in the development of hate speech detection systems. Our research questions are as follows: 
    
    \begin{enumerate}
        \item[\textsc{rq1}] Can we use open-source hate speech training data to monitor anti-LGBTQ+ hate speech in real world instances of social media? and;
        \item[\textsc{rq2}] How do the social, cultural, and linguistic contexts of open-source training data impact on the suitability of anti-LGBTQ+ hate speech detection?
    \end{enumerate}
    
    In order to address \textsc{rq1}, we compare and contrast two anti-LGBTQ+ hate speech detection systems. We provide an in depth description of the data sources in Section \ref{subsubsec:data_sources} and our system development pipeline in Section \ref{subsec:model_development}. Once we develop the detection systems, we apply the detection systems on real-world samples of social media data to monitor anti-LGBTQ+ hate speech across different geographic dialects.
    
    We opted for a mixed-methods approach to address this emergent area of enquiry. This is because \textsc{rq2} can only be addressed qualitatively as we consider the suitability of the detection systems and the sociocultural relevance of the predicted outputs. We will address \textsc{rq2} in the discussion (Section \ref{sec:discussion}); however, we have provided relevant sociolinguistic, cultural, and political information in Section \ref{subsubsec:communities_of_interest} to contextualise our discussion.

\subsection{Data Sources}
\label{subsubsec:data_sources}
    
    As part of our investigation, we use two open-source training data sets to develop our anti-LGBTQ+ hate speech detection systems in our investigation: \citet{ousidhoum_multilingual_2019} (\textit{Multilingual and Multi-Aspect Hate Speech Data Set}; \textsc{mlma}) and \citet{chakravarthi_dataset_2021} (\textsc{ltedi})\footnote{We refer to it as \textsc{ltedi} with reference to its central role in the various shared tasks hosted as part of the \textit{Language Technology for Equity, Diversity, and Inclusion}}. The \textsc{mlma} and \textsc{ltedi} were chosen due to the availability of data and documentation to understand the data set collection and annotation process. 

    \begin{table}[t]
        \centering
            \begin{tabular}{lccc}
            \hline \textbf{Class} & \textbf{\textsc{eng}} & \textbf{\textsc{tam}} & \textbf{\textsc{tam}-\textsc{eng}} \\ \hline
            \textsc{homo} & 276 & 723 & 465 \\
            \textsc{trans} & 13 & 233 & 184 \\
            \textsc{none} & 4,657 & 3,205 & 5,385 \\
            \textbf{Total} & 4,946 & 4,161 & 6,034 \\
            \hline
        \end{tabular}
        \caption{\label{tab:chakravathi} The class distribution of YouTube comments based on the three-class classification system (homophobic (\textsc{homo}), transphobic (\textsc{trans}), and non-anti-LGBTQ+ (\textsc{none}) content) by language condition (English (\textsc{eng}), Tamil (\textsc{tam}), and Tamil-English (\textsc{tam}-\textsc{eng})) in \citet{chakravarthi_dataset_2021}.}
    \end{table}
    
    The \textsc{mlma} is a multilingual hate speech data set for posts/tweets from X (Twitter) for English, French, and Arabic \cite{ousidhoum_multilingual_2019}. The authors took a keyword search approach by retrieving posts/tweets which matched a list of common slurs, controversial topics, and discourse patterns typically found in a hate speech. This approach proved challenging due to the high-rates of code-switching in the English and French conditions and Arabic diglossia. The posts/tweets were then posted on the crowd-sourcing platform, Mechanical Turk, for public annotation.
    
    One of the most well-documented anti-LGBTQ+ training data sets is the English, Tamil, and English-Tamil anti-LGBTQ+ hate speech data set developed by \citet{chakravarthi_dataset_2021}. The data set contains public comments to LGBTQ+ videos on YouTube. The comments were manually annotated based on a three-class (i.e., homophobic, transphobic, and non-anti-LGBTQ+ hate speech). The training data was tested with three language models: \textsc{muril} \cite{khanuja_muril_2021}, \textsc{mbert} \cite{pires_how_2019}, and \textsc{xlm-roberta} \cite{conneau_unsupervised_2020}. 
    
    The results show that transformer-based models, such as \textsc{bert}, outperformed statistical language models with minimal fine-tuning. The best performing \textsc{bert}-based system for English yielded an averaged $F_1$-score of 0.94 \cite{maimaitituoheti_ablimet_2022}. This anti-LGBTQ+ training data set has since expanded to a suite of additional language conditions such as Spanish \cite{garcia-diaz_umucorpusclassifier_2020}, Hindi and Malayalam \cite{kumaresan_homophobia_2023}, and Telugu, Kannada, Gujarati, Marathi, and Tulu \cite{chakravarthi_overview_2024}.
    
    We discuss the similarities and differences between the two data sets in relation sociocultural considerations regarding the data collection strategy in Section \ref{subsubsec:data_collection}, the annotation strategy in Section \ref{subsubsec:annotation_process}, and the cultural alignment in Section \ref{subsubsec:cultural_alignment} derived from available documentation.

\subsubsection{Data Collection} 
\label{subsubsec:data_collection}

    \begin{figure}[t]
        \centering
        \includegraphics[height=0.375\textwidth]{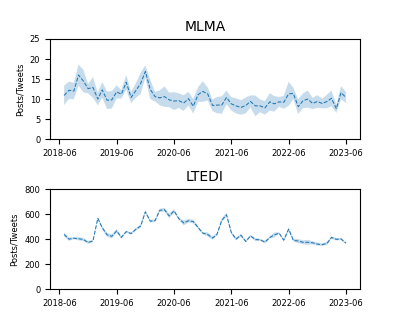}
        \caption{Model comparison of anti-LGBTQ+ hate speech on ten randomised samples of 10,000 posts/tweets per month from India between June 2018 to June 2023 including grouped mean and the upper and lower confidence intervals.}
        \label{fig:india}
    \end{figure}
    
    The developers of the \textsc{mlma} took a culturally-agnostic approach with limited information on the data collection points; however, evidence of code-switching between English with Hindi, Spanish, and French posed a challenge to annotators. The \textsc{mlma} took a keyword search approach to filter X (Twitter) for instances hate speech. The keywords in relation to anti-LGBTQ+ hate in English included: \textit{dyke}, \textit{twat}, and \textit{faggot}. This contrasts \textsc{ltedi} which took a content search approach of users reacting to LGBTQ+ content from India. 
    
    The high-level of code-switching and script-switching between English and other Indo-Aryan and Dravidian languages provides some level of social, cultural, and linguistic information of the training data. Both training data sets are comparable in size; however, \textsc{mlma} is 13.2\% larger than \textsc{ltedi} by number of observations. The proportion of anti-LGBTQ+ hate speech in the \textsc{mlma} is 9.1\% while the proportion of anti-LGBTQ+ hate speech in the \textsc{ltedi} is 5.8\%.
    
\subsubsection{Annotation Process} 
\label{subsubsec:annotation_process}

    \citet{bender_data_2018} proposed including data statement framework in the hope to mitigate different forms of social bias by dutifully documenting the NLP production process. Neither data sets provided annotator metadata  \cite{bender_data_2018}; therefore, we can only infer some of the annotator information from available documentation. Where the \textsc{mlma} took a crowd-sourcing approach, the \textsc{ltedi} data set were annotated by members of the LGBTQ+ communities. Based on the limited details, \textsc{ltedi} we know the annotators were English speakers based at the National University of Ireland Galway. Unsurprisingly, the \textsc{mlma} at 0.15 is lower than \textsc{ltedi} at 0.67 based on Krippendorf's alpha where 1 suggests perfect reliability while 0 suggests no reliability beyond chance.

\subsubsection{Cultural Alignment} 
\label{subsubsec:cultural_alignment}

    With limited documentation to the data set collection and annotation process beyond the system description papers, we tentatively determine the \textsc{ltedi} is largely in alignment with anti-LGBTQ+ discourse from the South Asian cultural sphere and the \textsc{mlma} as culturally-undetermined anti-LGBTQ+ rhetoric. This creates a useful contrast which not only compares the efficacy of two training data sets, but also anti-LGBTQ+ behaviour in different varieties of World Englishes which are influenced by their own unique social, cultural, and linguistic contexts \cite{kachru_other_1982}. We predict the data set collection and annotation approaches will have an impact on the outputs of the automatic detection systems.

    \begin{figure*}[t]
        \centering
        \includegraphics[height=0.3\textwidth]{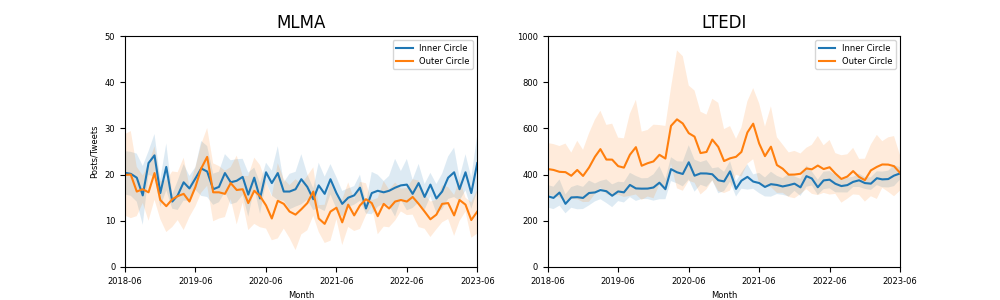}
        \caption{Comparison of anti-LGBTQ+ hate speech detected in 10,000 samples of posts/tweets from inner- and outer-circle varieties of English between June 2018 to June 2023 including grouped mean and the upper and lower confidence intervals.}
        \label{fig:compare}
    \end{figure*}
    
\subsection{System Development}
\label{subsec:model_development}

    \begin{table}
        \centering
        \begin{tabular}{lcccc}
        \hline
            \multirow{2}{*}{} &
          \multicolumn{2}{c}{\textbf{Macro}} &
          \multicolumn{2}{c}{\textbf{Weighted}} \\
          & \textit{Base} & \textit{Retrain} & \textit{Base} & \textit{Retrain} \\
            \hline
            \textsc{ltedi} & 0.78 & 0.81 & 0.95 & 0.96 \\
            \textsc{mlma} & 0.83 & 0.83 & 0.94 & 0.94 \\
            \hline
        \end{tabular}
        \caption{\label{tab:model_eval} Model evaluation metrics comparing the four candidate models by average macro $F_1$-score and average weighted $F_1$-score.}
    \end{table}
    
    The first phase of our investigation involves developing multiclass classification models to detect anti-LGBTQ+ hate speech in English. We opted for a transformer-based language modelling approach. Even though the focus of \textsc{ltedi} is YouTube, we can adapt Pretrained Language Models (\textsc{PLM}s) to specific domains, or register of language, through pretraining with additional samples of text \cite{gururangan_dont_2020}.

    We initially trained two classification models with minimal feature engineering in order to determine the best approaches to develop our automatic detection systems. We split the training data into training, development, and test sets with a train:development:test split of 90:5:5. We used Multi-Class Classification model from the Simple Transformers\footnote{https://simpletransformers.ai/} Python package to finetune and train the multi-class classification model. We trained each model for 8 iterations. We used AdamW as the optimiser \cite{loshchilov_decoupled_2018}. Our baseline \textsc{plm} is \textsc{xlm-roberta}, which is a cross-lingual transformer-based language model \cite{conneau_unsupervised_2020}. 

\subsubsection{Feature Engineering}
\label{subsubsec:feature_engineering}

    Class imbalance had an effect on our detection system. Therefore, we collapsed the multiple classes from each training data set into a binary classification. We also removed the confidentialised usernames and URLs from \citet{ousidhoum_multilingual_2019}, as we could not mask these high-frequency tokens from the classification model. We used RandomOverSampler from the Imbalanced Learn\footnote{https://imbalanced-learn.org/} Python package to upsample the minority classes. We address the register discrepancy in \citet{chakravarthi_dataset_2021}. We retrained \textsc{xlm-roberta} with 120,000 samples of X (Twitter) language data from the \textsc{cglu} \cite{dunn_mapping_2020}. The composition of the language data included 10,000 samples from each language condition.
    
\subsubsection{Model Evaluation}
\label{subsubsec:model_evaluation}

    We present the model evaluation metrics in Table \ref{tab:model_eval}. In Table \ref{tab:model_eval}, we compare the model evaluation results for the four candidate models (\textsc{ltedi}\textsubscript{\textsc{b}}, \textsc{ltedi}\textsubscript{\textsc{r}}, \textsc{mlma}\textsubscript{\textsc{b}}, and \textsc{mlma}\textsubscript{\textsc{r}}). The model performance improved in three of the four candidate models based on both macro average and weighted average $F_1$-score. Surprisingly, there were no differences between the two approaches for the \textsc{mlma} models. With a focus on the anti-LGBTQ+ class, domain adaptation improved the $F_1$-score from 0.58 to 0.64 for the \textsc{ltedi}\textsubscript{\textsc{r}} model. The $F_1$-score for the \textsc{mlma}\textsubscript{\textsc{r}} remains unchanged at 0.69. Based on the model performance metrics for the four candidate models, we advanced with the \textsc{ltedi}\textsubscript{\textsc{r}} and \textsc{mlma}\textsubscript{\textsc{r}} classification models with domain adaptation and feature engineering during finetuning. We continued to apply domain adaptation in both systems despite not seeing significant improvements in the \textsc{mlma}\textsubscript{\textsc{r}} model to maintain consistency between the two classification models.

    \begin{figure*}
        \centering
        \includegraphics[height=1.2\textwidth]{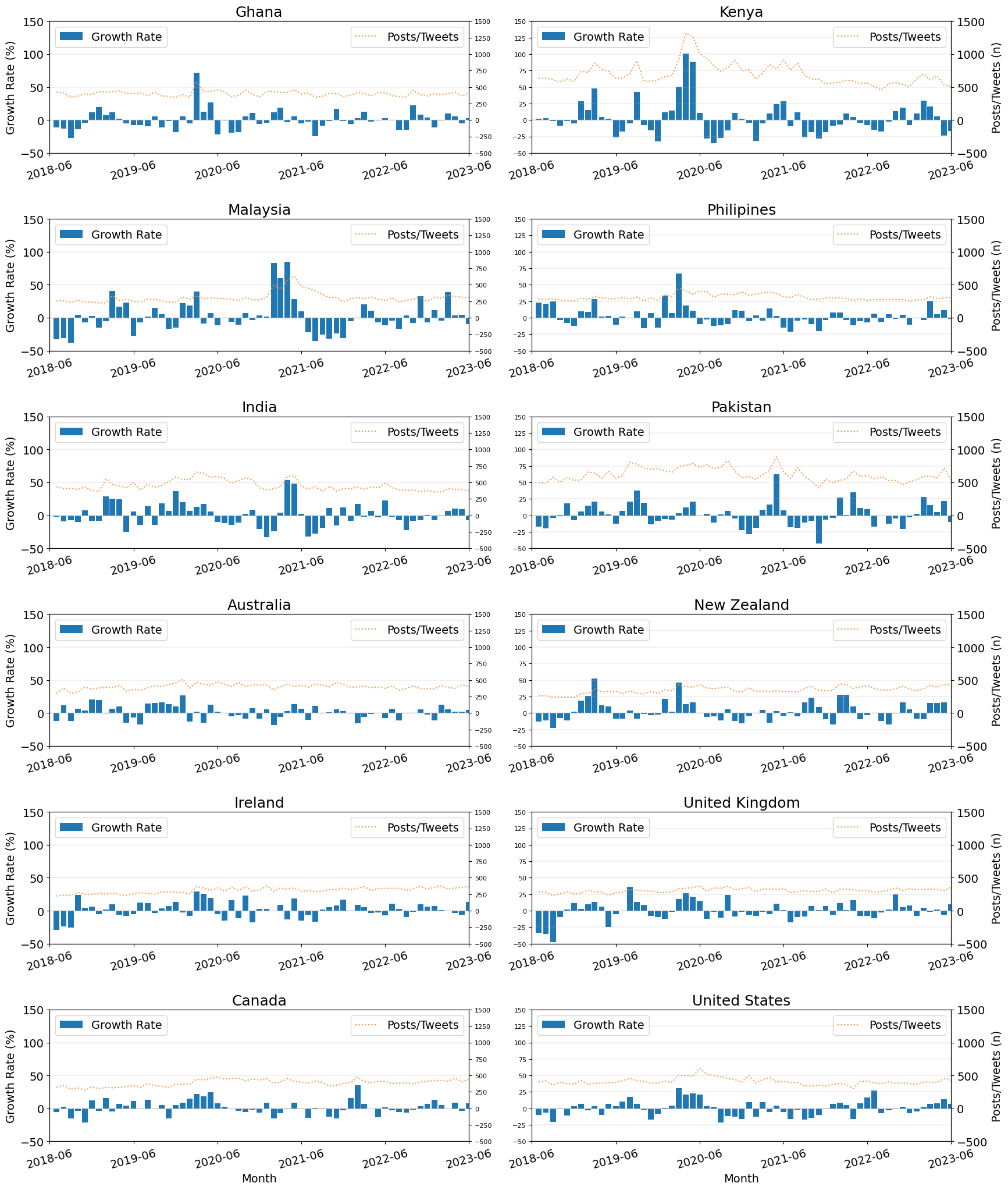}
        \caption{Quarterly growth rate of anti-LGBTQ+ hate speech detected with the \textsc{ltedi} model with number of posts/tweets by country between June 2018 and June 2023.}
        \label{fig:growth_ltedi}
    \end{figure*}

\subsection{Communities of Interest}
\label{subsubsec:communities_of_interest}

    Even though the \textsc{mlma} is supposedly culturally-agnostic, we have broadly identified the cultural alignment within the \textsc{ltedi} based on the data set collection and annotation process outlined in \citet{chakravarthi_dataset_2021}. More specifically, high-levels of code-switching and script-switching between English, Hindi, and Tamil in the \textsc{ltedi} suggests the presence of an Indian English substrate in the training data. Written English is often treated as homogeneous language; however, geographic-dialects represented by national-varieties of English maintain a constant-level of variation \cite{dunn_stability_2022}.
    
    Furthermore, the presence of Indian English on social media, or English spoken and written in India introduced as a result of British colonisation \cite{hickey_legacies_2005}, is uncontested \cite{rajee_analyzing_2024}. In the three concentric circles model of World Englishes, Indian English is categorised as an outer-circle variety of English \cite{kachru_standards_1985}. Outer-circle and inner-circle varieties of English are defined as national-varieties with British colonial ties. The distinguishing feature of outer-circle varieties is that English is not the primary language of social life and the government sector. These outer-circle varieties of English often co-exist alongside other indigenous languages.

    In order to test for the influence of social, cultural, and linguistic factors, we retrieved samples of social media language from outer-circle and inner-circle varieties of English. Outer-circle varieties of English as written English originating from Ghana, India, Kenya, Malaysia, the Philippines, and Pakistan. Similarly, inner-circle varieties as written English originating from Australia, Canada, Ireland, New Zealand, the United Kingdom, and the United States. The data source of our social media language data comes from a subset of \textsc{cglu} corpus which contains georeferenced posts/tweets from X (Twitter) \cite{dunn_mapping_2020}.

    For each national-variety of English, we filtered the data for tweets in English. All posts/tweets were processed with hyperlinks, emojis, and user identifying information removed. In addition to the monthly samples for each country, we re-sampled monthly tweets from India over ten iterations to determine the impact of our sampling methodology. All posts/tweets were produced between July 2018 to June 2023. Of relevance to our analysis, the countries associated with these national-varieties all criminalised same-sex sexual activity as a legacy of the English common law legal system (with the exception of the Philippines) \cite{han_british_2014}. All but four of these countries (Kenya, Ghana, Pakistan, Malaysia) have since decriminalised same-sex sexual activity. However, LGBTQ+ rights vary significantly between countries and LGBTQ+ communities continue to face discrimination in response to increased anti-LGBTQ+ legislation in the United States disproportionately affecting transgender people \cite{canady_mounting_2023}.

\section{Results}
\label{sec:results}

    We dedicate the current section to describe the results of the second phase of our investigation. This phase involved applying the candidate models to automatically detect anti-LGBTQ+ hate speech on real-world instances of social media data in English. Firstly, we applied both anti-LGBTQ+ hate speech detection models on the ten randomised monthly samples of social media language data from India using the same sampling methodology for other national-varieties of English. The results are shown in Figure \ref{fig:india}. As expected, the \textsc{ltedi}\textsubscript{\textsc{r}} model predicted higher rates of anti-LGBTQ+ hate speech; however, what was unexpected were the low number of predictions from the \textsc{mlma}\textsubscript{\textsc{r}} model. The narrow confidence intervals suggest little instability between the different samples and the predictions remained constant across samples.

    After validating our sampling methodology by visually inspecting the ten randomised monthly samples from India, we applied both models on random samples of inner- and outer-circle varieties of English. We compared the results of the detection models as visualised in Figure \ref{fig:compare}. These were consistent with our initial results. The rate of anti-LGBTQ+ hate speech remained constant according to the \textsc{mlma}\textsubscript{\textsc{r}} model, while anti-LGBTQ+ hate speech has increased over time based on a visual inspection of the results. Of interest to our investigation, the \textsc{mlma}\textsubscript{\textsc{r}} model identified a higher proportion of anti-LGBTQ+ hate speech in inner-circle varieties of English. We saw an inverse relationship with the \textsc{ltedi}\textsubscript{\textsc{r}} where we see a higher proportion of anti-LGBTQ+ hate speech in outer-circle varieties of English. The wide confidence intervals of the \textsc{ltedi}\textsubscript{\textsc{r}} suggests greater between-variety instability in outer-circle varieties of English.

    We calculated the quarterly growth rates for each variety of English for the predictions from the \textsc{ltedi}\textsubscript{\textsc{r}}. We included the total number of predicted posts/tweets in our visualisation as shown in Figure \ref{fig:growth_ltedi}. The growth rates allowed us to determine the growth rate for each variety of English independently. The results suggest the growth rate of predicted anti-LGBTQ+ hate speech has remained stable over time. 

    \begin{figure}[t]
        \centering
        \begin{minipage}[b]{0.5\textwidth}
            \begin{subfigure}{\textwidth}
                \centering
                \subcaption{Training Data}
                \includegraphics[height=0.45\linewidth]{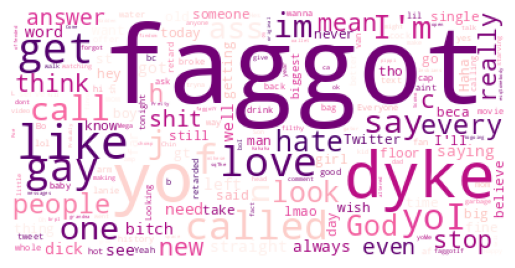}
            \end{subfigure}\par
            \begin{subfigure}{\textwidth}
                \centering
                \subcaption{Predicted Outputs}
                \includegraphics[height=0.45\linewidth]{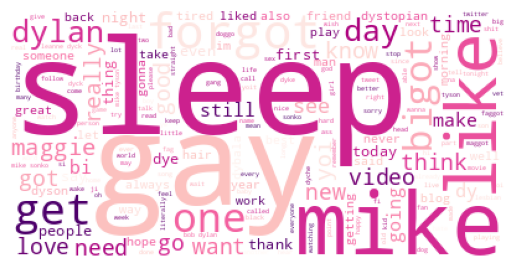}
            \end{subfigure}
        \end{minipage}
        \caption{\textsc{mlma} Wordcloud.}
        \label{fig:wordcloud-mlma}
    \end{figure}

    \begin{figure}[t]
        \centering
        \begin{minipage}[b]{0.5\textwidth}
            \begin{subfigure}{\textwidth}
                \centering
                \subcaption{Training Data}
                \includegraphics[height=0.45\linewidth]{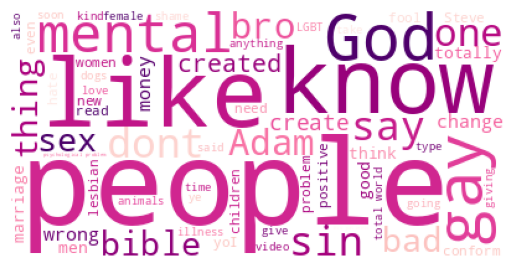}
            \end{subfigure}\par
            \begin{subfigure}{\textwidth}
                \centering
                \subcaption{Predicted Outputs}
                \includegraphics[height=0.45\linewidth]{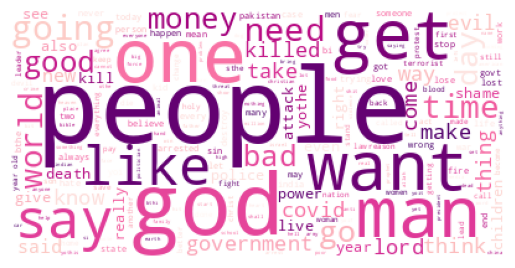}
            \end{subfigure}
        \end{minipage}
        \caption{\textsc{ltedi} Wordcloud.}
        \label{fig:wordcloud-ltedi}
    \end{figure}

\section{Discussion}
\label{sec:discussion}

    The results from our study raises some interesting questions on the efficacy of these systems on real-world instances of social media data. With regards to the first research question, our transformer-based multiclass classification model enabled us to detect instances of anti-LGBTQ+ hate speech from samples of georeferenced posts/tweets from X (Twitter). We were able to manipulate the predicted outputs into different forms of time series as shown in Figures \ref{fig:compare} and \ref{fig:growth_ltedi}. The level of anti-LGBTQ+ hate speech has maintained a constant rate of growth despite decreasing usership on the social media platform since the acquisition of X (Twitter) by Elon Musk in 2022. The results suggest that anti-LGBTQ+ hate speech on X (Twitter) is indeed increasing in both rate and volume over time \cite{hattotuwa_transgressive_2023}.
    
    When we compare the predicted results between the \textsc{mlma}\textsubscript{\textsc{r}} and the \textsc{ltedi} models, we can see significant differences between the two models. This is particularly obvious when we compare the predicted outputs in Figures \ref{fig:india} and \ref{fig:compare}, where the \textsc{ltedi}\textsubscript{\textsc{r}} model on average predicted 50 times more instances of anti-LGBTQ+ hate speech than the \textsc{mlma}\textsubscript{\textsc{r}}. These was unexpected as the model evaluation metrics during model development suggested the \textsc{mlma}\textsubscript{\textsc{r}} model performed marginally better than the \textsc{ltedi}\textsubscript{\textsc{r}}. Considering both the sampling methodology and the model development approaches were held constant between the models, we propose the differences we see in the predicted outputs is a result of the open-source training data.

    One challenge of applying multiclass classification models on unknown data is that there is no simple method to validate the results. This is because we do not have access to labelled training, development, and test sets to evaluate the model performance. We are therefore reliant on qualitative methods to validate the performance of our detection models. Figure \ref{fig:wordcloud-mlma} is a visual representation of the word-token frequencies between the open-source training data (a) and the predicted anti-LGBTQ+ hate speech (b) from the samples of posts/tweets. The most prominent word-token in the training data is \textit{faggot} followed by \textit{dyke}. This is not unexpected as these word-tokens (including \textit{twat}) were used to identify instances of anti-LGBTQ+ hate speech on X (Twitter). Counterintuitively, we did not see a similar distribution in the predicted outputs.
    
    With reference to Figure \ref{fig:wordcloud-mlma}, the word-tokens with the highest frequency in the predicted output were not \textit{faggot} or \textit{dyke}, but \textit{sleep} and \textit{gay}. When we filtered for the keyword search terms in the samples, we found few instances across the varieties of English as shown in Tables \ref{tab:outer} and \ref{tab:inner}. This is unexpected as the keyword search terms are highly prevalent in inner-circle varieties of spoken English (such as the United Kingdom and Ireland) \cite{love_swearing_2021}. This is supported by the higher word-token frequencies in inner-circle varieties of English as shown in Tables \ref{tab:outer} and \ref{tab:inner}. We attribute the infrequent occurrence of LGBTQ+ slurs in direct response to X (Twitter) rules which discourages hateful conduct on the platform. 
    
    Our analysis of the \textsc{mlma}\textsubscript{\textsc{r}} model suggests a relationship between the training data and the resulting detection model. Incidentally, we also observe this bias towards inner-circle varieties of English in Figure \ref{fig:compare} where the \textsc{mlma}\textsubscript{\textsc{r}} is more inclined to identify more anti-LGBTQ+ hate speech in inner-circle than outer-circle varieties of English. This leads our discussion to the second research question where we determine how the social, cultural, and linguistic context impacts the efficacy of anti-LGBTQ+ hate speech detection. Although anti-LGBTQ+ discourse is consistent across languages \cite{locatelli_cross-lingual_2023}, slurs and swearwords are not \cite{jay_pragmatics_2008}. This form of cultural bias toward inner-circle varieties of English (or oversight of outer-circle varieties) introduced during the data collection process, raises questions on the suitability of the \textsc{mlma}\textsubscript{\textsc{r}} model in monitoring anti-LGBTQ+ hate speech. 

    \begin{table}[t]
        \centering
            \begin{tabular}{ccccc}
            \hline \textbf{Variety} & \textbf{\textit{dyke}} & \textbf{\textit{faggot}} & \textbf{\textit{twat}}\\ \hline
            \textsc{gh} & 8 &  2 & 6 \\
            \textsc{in} & 5 &  - & 6 \\
            \textsc{ke} & 1 &  4 & 7 \\
            \textsc{my} & 3 &  4 & 14 \\
            \textsc{ph} & 8 &  4 & 8 \\
            \textsc{pk} & 3 &  6 & 6 \\
            \hline
        \end{tabular}
        \begin{tabular}{c}
            \hline \textbf{\textit{gay}} \\ \hline
            353 \\
            226 \\
            295 \\
            500 \\
            701 \\
            478 \\
            \hline
        \end{tabular}
        \caption{\label{tab:outer} Frequency of LGBTQ+ related slurs for outer circle varieties of English.}
    \end{table}

    \begin{table}[t]
        \centering
        \begin{tabular}{ccccc}
            \hline \textbf{Variety} & \textbf{\textit{dyke}} & \textbf{\textit{faggot}} & \textbf{\textit{twat}}\\ \hline
            \textsc{au} & 5 & 12 & 48 \\
            \textsc{ca} & 16 & 13 & 19 \\
            \textsc{ie} & 15 & 16 & 62 \\
            \textsc{nz} & 6 & 14 & 53 \\
            \textsc{uk} & 23 &  9 & 148 \\
            \textsc{us} & 19 & 11 & 13 \\
            \hline
        \end{tabular}
        \begin{tabular}{c}
            \hline \textbf{\textit{gay}} \\ \hline
            635 \\
            623 \\
            659 \\
            627 \\
            679 \\
            875 \\
            \hline
        \end{tabular}
        \caption{\label{tab:inner} Frequency of LGBTQ+ related slurs for inner circle varieties of English.}
    \end{table}

    As we determined the \textsc{ltedi}\textsubscript{\textsc{r}} model to be more culturally aligned with the South Asian context, we initially predicted the \textsc{ltedi} model would be more appropriate for South Asian contexts. However, the results suggest the \textsc{ltedi}\textsubscript{\textsc{r}} model as more fit for purpose in contrast to the \textsc{mlma}\textsubscript{\textsc{r}} model. Not only do we observe high-congruency between the \textsc{ltedi}\textsubscript{\textsc{r}} model output and the outer-circle varieties of English as shown in Figure \ref{fig:compare}, the word-token frequencies between the training data (a) and the predicted outputs (b) in the \textsc{ltedi} appear to have a similar distribution as shown in Figure \ref{fig:wordcloud-ltedi}. 

    Curiously, both the training data and predicted output lack slurs. Instead, we see word-tokens associated with community (e.g., \textit{people}) and religion (e.g., \textit{bible}, \textit{god}, and \textit{Adam} possibly in reference to the Abrahamic creation myth of \textit{Adam and Eve}). This is unsurprising as anti-LGBTQ+ legislation is often rooted in puritanical beliefs on morality \cite{han_british_2014}. With reference to Figure \ref{fig:growth_ltedi}, we observed a possible link between the increased growth rate with nationwide response to the Covid-19 pandemic. Once again this raises a question on the validity of the predicted outputs and whether the posts/tweets are anti-LGBTQ+ or religious/spiritual in nature (or indeed, both).

\section{Conclusion}
\label{sec:conclusion}

    The findings from this current paper raises a number challenges in applying hate speech detection in a real-world context. Even within national-varieties of English, we observed the impacts of social, cultural, and linguistic factors. For example, the \textsc{ltedi}\textsubscript{\textsc{r}} which was culturally aligned with Indian English was more sensitive to outer circle varieties of English, while the \textsc{mlma}\textsubscript{\textsc{r}} model was slightly more sensitive to inner circle varieties of English. We conclude that monitoring anti-LGBTQ+ hate speech with open-source training data is not problematic in itself; however, we must interpret these empirical outputs with qualitative insights to ensure these systems are fit for purpose.

\section*{Ethics Statement}
\label{sec:ethics_statement}

    The purpose of this paper is to investigate the suitability of using open-source training data to develop a multiclass classification model to monitor and forecast levels of anti-LGBTQ+ hate speech on social media across different geographic dialect contexts in English. This study contributes to the efforts in mitigating harmful hate speech experienced by LGBTQ+ communities. In our investigation, we combine methods from NLP, sociolinguistics, and discourse analysis to evaluate the effectiveness of anti-LGBTQ+ hate speech detection.

    We recognise the importance of advocate and activist-led research in particular by members of under-represented and minoritised communities \cite{hale_engaging_2008}. The lead author acknowledges their positionality as an active advocate and a member of the LGBTQ+ community \cite{wong_queer_2023}. The lead author is familiar with anti-LGBTQ+ discourse both in online and offline spaces and its harmful effects on members of the LGBTQ+ communities.

    As discussed in Section \ref{sec:conclusion}, we support the critique of \citet{parker_is_2023} for NLP researchers to reflect on the efficacy and suitability of hate speech detection models. The development of hate speech data sets impose a `diversity tax' on already marginalised LGBTQ+ communities. Originally coined by \citet{padilla_ethnic_1994}, this refers to the unintentional burden placed on marginalised peoples to address inequities, exclusion, and inaccessibility particularly in a research context. NLP researchers need to work alongside key-stakeholders (e.g., affected communities, advocates, and activists) as well as social media platforms, non-profit organisations, and government entities to determine the solutions of this social issue.

    The inclusion of unobfuscated examples of slurs, hate speech, and offensive language towards LGBTQ+ communities is a deliberate attempt to initiate the process of reclaiming and re-appropriating some anti-LGBTQ+ slurs in NLP research. Currently, there are limited best practice guidelines on the obfuscation of profanities in NLP research \cite{nozza_state_2023}. \citet{worthen_queers_2020} theorised that anti-LGBTQ+ slurs are used to stigmatise violations of social norms. Re-appropriating these stigmatising labels can enhance what were once devalued social identities \cite{galinsky_reappropriation_2003}. This process of `cleaning' and `detoxifying' slurs is also a process of resistance and to reclaim power and control \cite{popa-wyatt_reclamation_2020}.

    We argue that within context of social media research giving unwarranted attention to slurs ignores the root of this social issue: hate speech expresses hate \cite{marques_expression_2023}. Many social media platforms have already put in place procedures to censor sensitive word-tokens; however, social media users continue to adopt innovative linguistic strategies such as \textit{voldermorting} \cite{van_der_nagel_networks_2018} and \textit{Algospeak} \cite{steen_you_2023} to contravene well-meaning moderation and censorship algorithms. Our results suggest hate speech training data sets do not identify the full breadth of hateful content on social media.
        
    This paper does not include human or animal participants. Furthermore, we abide by the data sharing rules of X (Twitter) and posts/tweets with identifiable personal details will not be shared publicly. The authors have no conflicts of interests to declare.

\section*{Limitations}

    In this section, we address some of the known limitations of our approach in addition to limitations of the open-source training data and the social media data we have used in the current study.

    \paragraph{Invisibility of Q+ identities} This paper uses the LGBTQ+ acronym to signify diverse gender and sexualities who continue to experience forms of discrimination and stigmatisation (namely Lesbian, Gay, Bisexual, and Transgender people). While the Q+ refers to those who are not straight or not cisgender (Queer+), we acknowledge the invisibility of other minorities who are often excluded from NLP research including intersex and indigenous expressions of gender, sexualities, and sex characteristics at birth.

    \paragraph{Sociocultural bias during data collection} Despite including more training data, the \textsc{mlma} identified significantly fewer instances of anti-LGBTQ+ hate speech than the \textsc{ltedi} across the national-varieties of English. With reference to the wordclouds produced from the training data for \textsc{mlma} and \textsc{ltedi} as shown in Figures \ref{fig:wordcloud-mlma} and \ref{fig:wordcloud-ltedi}, there is a high likelihood the keyword search (on \textit{dyke}, \textit{twat}, and \textit{faggot}) during the data collection process has caused the classification model to over-fit the training data. Similarly, the religious subtext in the \textsc{ltedi} training data reinforces polarising beliefs that religion is anti-LGBTQ+. Furthermore, these detection systems do not account for semantic bleaching or the reclamation of slurs \cite{popa-wyatt_reclamation_2020}.

    \paragraph{Pitfalls of large language models} We acknowledge the cultural and linguistic biases introduced through the \textsc{plm}s used in our transformer-based approach. However, we have mitigated some of these impacts through domain adaptation \cite{liu_roberta_2019}. With reference to Figure \ref{fig:wordcloud-mlma}, we have reason to believe the transformer-based detection systems erroneously classified \textit{dylan}, \textit{mike} and \textit{like} with \textit{dyke}. A breakdown of the character-trigrams (\textsc{\#dy}, \textsc{dyk}, \textsc{yke}, and \textsc{\#ke}) confirms this belief.

    \paragraph{Class imbalance and distribution} We were able to improve the performance of the detection model during model development by up-sampling the minority classes. The \textsc{ltedi} detected a constant proportion of anti-LGBTQ+ hate speech between 5-10\% for all varieties of English which is a similar proportion of anti-LGBTQ+ hate speech in the training data (or 5.8\% of the training data). This raises potential questions on the efficacy of transformer-based classification models.

    \paragraph{Further work} We welcome NLP researchers to address these limitations in their research especially on increasing the visibility of Q+ communities and the sociocultural biases shown in open-source training data sets and large language models.

\section*{Acknowledgements}

    The lead author wants to thank Dr. Benjamin Adams (University of Canterbury | Te Whare Wānanga o Waitaha) and Dr. Jonathan Dunn (University of Illinois Urbana-Champaign) for their feedback on the initial manuscript. The lead author wants to thank the three anonymous peer reviewers and the programme chairs for their constructive feedback. Lastly, the lead author wants to thank Fulbright New Zealand | Te Tūāpapa Mātauranga o Aotearoa me Amerika and their partnership with the Ministry of Business, Innovation, and Employment | Hīkina Whakatutuki for their support through the Fulbright New Zealand Science and Innovation Graduate Award. 

\bibliography{references.bib}

\end{document}